\def\year{2022}\relax
\documentclass[letterpaper]{article} 
\usepackage{aaai22}  
\usepackage{times}  
\usepackage{helvet}  
\usepackage{courier}  
\usepackage[hyphens]{url}  
\usepackage{graphicx} 
\usepackage{natbib}  
\usepackage{caption} 
\DeclareCaptionStyle{ruled}{labelfont=normalfont,labelsep=colon,strut=off} 
\usepackage{algorithm}
\usepackage{algorithmic}

\usepackage{newfloat}
\usepackage{listings}
\usepackage{subfigure}
\usepackage{booktabs}
\usepackage{multirow}
\usepackage{amsfonts,amssymb,amsmath}
\floatstyle{ruled}
\newfloat{listing}{tb}{lst}{}
\floatname{listing}{Listing}
\title{Incomplete Multi-view Clustering via Cross-view Relation Transfer}
\author{
    Yiming~Wang,~Dongxia~Chang,~Zhiqiang~Fu, and~Yao~Zhao
}
\affiliations{

}

\usepackage{bibentry}

\begin{document}

\maketitle

\begin{abstract}


In this paper, we consider the problem of multi-view clustering on incomplete views. Compared with complete multi-view clustering, the view-missing problem increases the difficulty of learning common representations from different views. To address the challenge, we propose a novel incomplete multi-view clustering framework, which incorporates cross-view relation transfer and multi-view fusion learning. Specifically, based on the consistency existing in multi-view data, we devise a cross-view relation transfer-based completion module, which transfers known similar inter-instance relationships to the missing view and recovers the missing data via graph networks based on the transferred relationship graph. Then the view-specific encoders are designed to extract the recovered multi-view data, and an attention-based fusion layer is introduced to obtain the common representation. Moreover, to reduce the impact of the error caused by the inconsistency between views and obtain a better clustering structure, a joint clustering layer is introduced to optimize recovery and clustering simultaneously. Extensive experiments conducted on several real datasets demonstrate the effectiveness of the proposed method.

\end{abstract}

\begin{figure*}[t]
	\centering
	\includegraphics[width=16.5cm]{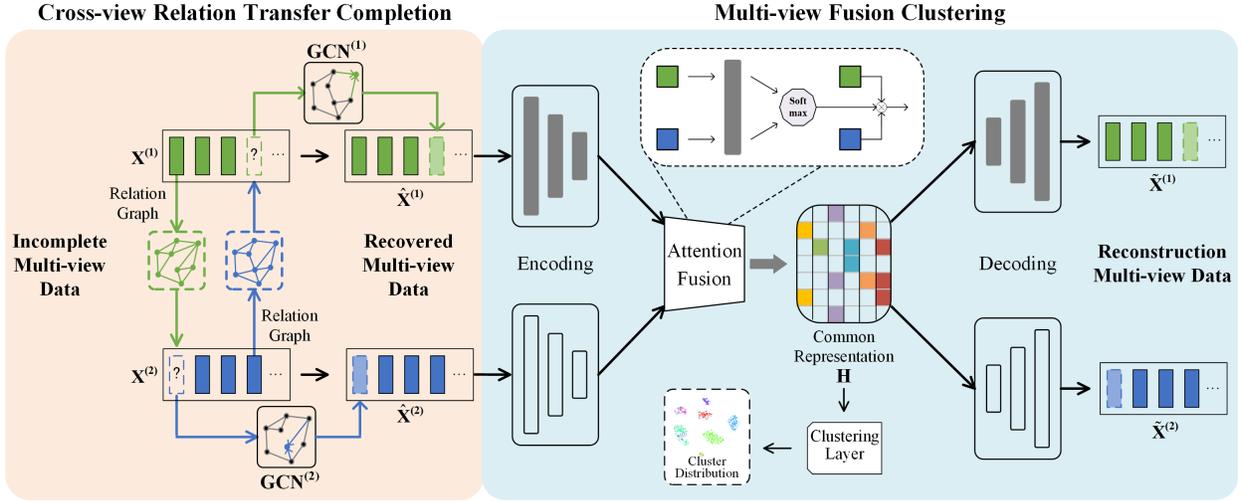}
	\caption{The framework of the proposed CRTC. In the figure, two-view data is used as a showcase. As shown, it consists of two main components: cross-view relation transfer completion and multi-view fusion clustering. Specifically, the cross-view relation transfer completion module recovers missing data by transferring similar inter-instances relationships of existing views. And the multi-view fusion clustering network adaptively fuses the features of each view to obtain a consistent representation suitable for clustering.}
\end{figure*}

\section{Introduction}
\noindent With the advance of data acquisition devices, real data often appears in multiple modalities or comes from multiple sources \cite{conf/colt/BlumM98}. Multi-view Clustering (MVC), as one of the most important unsupervised multi-view learning tasks, has attracted lots of attention. The goal of MVC is to improve clustering performance by integrating features from multiply views. Towards this goal, a variety of related methods have been proposed \cite{journals/pami/YuTLGSMM12,conf/aaai/LiNHH15,journals/pami/LiuZLWTYSWG19}. Moreover, MVC is widely applied in various applications, such as object segmentation, information retrieval, and saliency detection. A common assumption for most MVC methods is that all of the views are complete. However, in real-world applications, some views often suffer from data missing, which makes most multi-view clustering methods inevitably degenerate. So how to reduce the impact of missing views and learn a consistent representation are challenging problems of incomplete multi-view learning.

To solve the above challenging problem, many incomplete multi-view clustering (IMC) methods have been proposed in the last few decades. These methods can be divided into three main categories: Non-negative Matrix Factorization (NMF) based methods, kernel based methods, and Deep Neural Networks (DNNs) based methods. NMF based methods \cite{conf/aaai/ZhaoDF17,conf/ijcai/HuC18} learn a common latent space for complete instances and private latent representations for incomplete instances. For example, Li et al. \cite{aaai/LiJZ14} first proposed NMF based partial multi-View clustering method to learn a latent subspace over two views, which achieves notable clustering performance. Motivated by these methods, a lot of partial multi-view clustering approaches \cite{journals/tnn/LiuJLZL15} extend NMF to data with more views. Kernel based methods \cite{conf/icdm/ShaoSY13,journals/pami/LiuZLWTYSWG19} leverage kernel matrices of the complete views for completing the kernel matrix of incomplete view. For example, MVL-IV \cite{journals/tip/XuT015} proposes an algorithm to accomplish multi-view learning with incomplete views by exploiting the connections of multiple views. These two types of methods show satisfactory performance, but suffer from two main disadvantages: (a) It is inefficient to employ them for large-scale datasets. (b) They need high time consumption. The third kind of methods is based on DNNs. Recently, DNNs have shown amazing power in handling multi-view data. Many recent studies \cite{conf/ijcai/WenZ0ZFX20,9258396} employ DNNs to solve IMC challenges, showing noticeable improvement on clustering performance. These DNNs based methods typically employ Autoencoders (AEs) to learn the common representations for clustering and employ Generative Adversarial Networks (GANs) to generate missing data. However, GAN-based methods have difficulty avoiding the problem of training instability. Furthermore, few methods consider cross-view consistency. 

Different from existing deep incomplete multi-view clustering approaches, we focus on recovering missing data based on the similar relationships between instances in the existing views. Considering consistency of multiple views, similarity relationships between instances in the existing views have availability in the missing view. On this basis, we design a novel completion model based on cross-view relation transfer and proposed a joint incomplete multi-view clustering framework named Cross-view Relation Transfer Clustering (CRTC). The main contributions of our method are summarized as follows:



\begin{itemize}
    \item  We devised a cross-view relation transfer completion module based on the consistency existing in multi-view data. To the best of our knowledge, our method is the first approach to recover missing data by transferring similarity relationships between instances in the existing views.
    \item Inspired by transformer technology, we introduce an attention-based fusion layer to obtain the common representation, which can efficiently reduce the negative influence of missing views.
    \item Missing data completion and multi-view clustering are incorporated to obtain the final clustering assignments. Moreover, extensive experiments show that CRTC makes a considerable improvement over the state-of-the-art methods.
\end{itemize}

\section{Related Work}
\textbf{Deep Incomplete Multi-view Clustering.}
With the development of deep learning, a number of recent studies \cite{9258396, lin2021completer} employ DNNs to solve incomplete multi-view clustering challenges, showing noticeable improvement on clustering performance. Based on producing missing data or not, most existing deep incomplete multi-view clustering methods could be roughly classified into two categories, \emph{i.e.}, learning a uniform representation without generating missing data \cite{conf/mm/0001ZZWFXZ20,conf/ijcai/WenZ0ZFX20,journals/tip/WangLSGJ21} and generating missing instances using adversarial learning \cite{journals/tip/WangDTGF21,conf/icdm/WangDTG018,9258396,9382955}. The first class of methods reduces the negative influence of missing views by modifying the multi-view fusion representation learning model. For example, CDIMC-net \cite{conf/ijcai/WenZ0ZFX20} introduces a self-paced strategy to select the most confident samples for model training to reduce the negative influence of outliers. Moreover, DIMC-net \cite{conf/mm/0001ZZWFXZ20} further develops a weighted fusion layer to obtain the common representation shared by all views. Instead, iCmSC \cite{journals/tip/WangLSGJ21} employs subspace learning to solve IMC, and maximize the intrinsic correlations among different modalities by deep canonical correlation analysis (DCCA) \cite{conf/icml/AndrewABL13}. Adversarial learning-based methods for missing completions typically use discriminators to optimize the generation process of missing instances. For example, CPM-Nets \cite{9258396} uses encoding networks to generate missing data and introduces an adversarial strategy to enforce the generated data for missing ones in each view to obey distribution of the observed data. GP-MVC \cite{journals/tip/WangDTGF21} introduces view-specific GAN with multi-view cycle consistency to generate the missing data of one view conditioning on the shared representation given by other views. Unlike these two types of methods, COMPLETER \cite{lin2021completer} generates the representations of the missing views by minimizing the conditional entropy of different views by dual prediction. Different from the current methods, we focus on generating missing data based on the similarity relationships of corresponding instances in the existing views.

\textbf{Graph Neural Networks.}
To mine non-Euclidean graph information, researchers have devised graph neural networks (GNNs) \cite{conf/iclr/KipfW17,conf/iclr/VelickovicCCRLB18} capable of encoding both graph structure and node characteristics for node latent representation. GNNs have successfully expanded deep learning techniques to non-Euclidean graph data with remarkable achievement made in multiple graph tasks, such as graph classification \cite{conf/nips/LiC0T20} and visual question answering \cite{conf/iccv/LiGCL19}. Additionally, thanks to the properties of graph convolution, many GNN-based graph clustering methods \cite{conf/ijcai/WangPHLJZ19,conf/www/FanWSLLW20} have been proposed. These methods typically construct adjacency matrices based on the K-nearest-neighbor (KNN) graph \cite{conf/www/Bo0SZL020} or attributed graph \cite{conf/ijcai/ChengWTXG20}, and learn graph embedding using graph autoencoder (GAE) \cite{journals/corr/KipfW16a}.  In our model, GNN is not directly used to learn common representations but to generate missing instances through inter-instances relationships.

\section{The proposed CRTC}

\subsection{Motivation and Notations}
As aforementioned analysis, existing incomplete multi-view clustering methods can be roughly grouped into three categories, \emph{i.e.}, NMF based methods, kernel based methods, and DNN based methods. Among them, methods of inferring missing data by NMF and kernel may suffer problems such as high time consumption. Moreover, most DNN based methods recover missing data using GAN, which may face training problems such as difficult convergence and gradient explosion. Meanwhile, similar inter-instance relationships in the existing view are hardly utilized in current multi-view clustering methods. From this point, we propose Cross-view Relation Transfer Clustering (CRTC), which performs multi-view recovery and clustering simultaneously. As illustrated in Fig.~1, CRTC consists of two main components: cross-view relation transfer completion and multi-view fusion clustering, which are jointly trained to recover missing data and obtain clustering assignments.

In the incomplete multi-view clustering, instances are characterized by multiple views and some views may be destroyed. Formally, given a multi-view incomplete dataset $\mathcal{X} = \{X^{(1)},...,X^{(V)}\}$ consisting of $N$ instances of $V$ views. And $X^{(v)} = \{x^{(v)}_1,...,x^{(v)}_N\} \in \mathbb{R}^{ {D_v} \times N}$ denotes the feature matrix in the $v$-th view, where $D_v$ is the feature dimension of $v$-th view. For convenience, matrix $M \in \mathbb{R}^{ N\times V}$ is used to record the view available and missing information, where $M_{i,v} = 0$ means the $i$-th instance is available in the $v$-th view, otherwise $M_{i,v} = 1$. And the goal of CRTC is to group all the $N$ instances into $C$ clustering.

\subsection{Cross-view Relation Transfer Completion}

In the task of incomplete multi-view clustering, recovering missing data has been a challenge and has a large impact on the clustering performance. Existing DNN based methods typically obtain the missing data using GAN, which fail to consider inter-instance relationship in the existing views and may suffer from training difficulties. To this end, we devise a cross-view relation transfer completion module, which transfers the known similar inter-instance relationships in the existing views to recover missing views.

\begin{figure}[t]
	\centering
	\includegraphics[width=8cm]{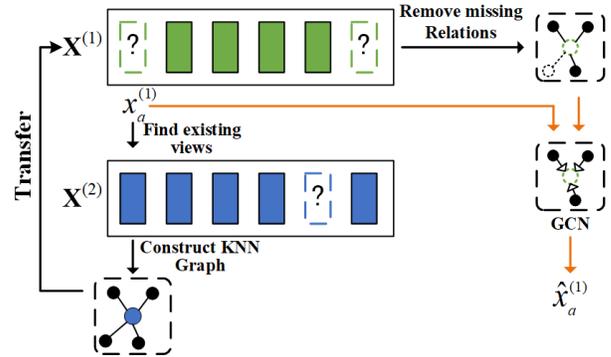}
	\caption{Illustration of cross-view relation transfer completion module}
\end{figure} 

Here we use the two-view data shown in Fig.~2 to show the process of the cross-view relation transfer completion. For the missing data $x^{(1)}_a$ in the first view,  the corresponding existing instance in the second view can be denoted as $x^{(2)}_a$. Since IMC is an unsupervised task, to obtain the existing instances related to $x^{(2)}_a$ in the second view, KNN graph is constructed based on the distance metric. And the $K$ neighbors of $x^{(2)}_a$ can be denoted by $K^{(2)}_a = \{x^{(2)}_{a^1},...,x^{(2)}_{a^K}\}$. Considering the consistency of multiple views, similarity relationships between instances in existing views are valid for the missing views. Therefore, we can transfer the KNN graph to the first view to find similar instances to the missing data $x^{(1)}_a$. Since some of instances in $K^{(2)}_a$ are missing in the first view, we remove the missing instances and obtain the transferred KNN graph $K^{(1)}_a = \{x^{(1)}_{a^1},...,x^{(1)}_{a^{K'}}\}$, where $K'\leq K$. In this way, we can obtain the transferred KNN graph of all missing data. For datasets with more than two views, there may be more known instances for missing data $x^{(1)}_a$, which leads to more transferred KNN graphs. We combine these graphs into a single graph to infer the missing data.

To recover the missing data, we apply a graph convolutional network (GCN) to infer the missing data based on the transferred relations. For convenience, the recovery of $x^{(1)}_a$ is presented as an example to introduce the completion process, and the input of GCN is the KNN graph $K^{(1)}_a$ consisting of features and edges. Then the recovered feature can be obtained by:
\begin{equation}
\hat{x}_a^{(1)} = \sigma(b + \sum_{j\in K^{(1)}_a}\frac{1}{|K^{(1)}_a|}w x_{a_j}^{(1)})
\end{equation}
where $\sigma(\cdot)$ is an activation function, $b$ is the bias, $w$ is the parameter matrix, and $x_{a_j}^{(1)}$ denotes the features of instances in $K^{(1)}_a$. Since the feature of $x^{(1)}_a$ is ‘NaN’, we remove the self-loop, which is a typical setting in GNN. 

Then, in order to recover all missing data according to corresponding transferred relations and existing instances, we minimize the reconstruction loss as
\begin{equation}
    L_{cr} = \sum_{i=1}^N\sum_{v=1}^V\sum_{j\in K^{(i)}_a} ||\hat{x}_i^{(v)}-x_{i_j}^{(v)}||^2 M_{i,v}
\end{equation}

\subsection{Multi-view Fusion Clustering}

To obtain the final clustering assignments, we develop a multi-view fusion clustering network, which mainly contains three parts: view-specific encoding and decoding, an attention fusion layer, and a clustering layer. In fact, in addition to the consistency existing in multi-view data, each view is also complementary and contains distinctive features that other views do not have. 
Therefore, to capture the view-specific features of different views, we use view-specific encoders to embed original heterogeneous data to the same size features. And the $i$-th embedding features of $v$-th view can be obtained as 
\begin{equation}
h_i^{(v)} = g^{(v)}(x_i^{(v)})
\end{equation}
where $g^{(v)}$ denotes the encoder for the $v$-th view. 
In order to flexibly integrate the view-specific features and mitigate the effects of missing data, we devise an attention fusion layer to obtain a common representation based on the learned view-specific features. Specifically, for a view-specific features $h^{(v)}_i$, we calculate the attention distribution $\alpha^{(v)}_i$ as 
\begin{equation}
\begin{aligned}
\alpha_i^{(v)} = & {\rm softmax}(s(w_i^{(v)}h_i^{(v)}, h_i^+)) \\
               = & \frac{exp(s(w_i^{(v)}h_i^{(v)}, h_i^+))}{\sum_{j=1}^V exp(s(w_i^{(j)}h_i^{(j)}, h_i^+))}\\
\end{aligned}
\end{equation}
where $w_i^{(v)}$ is the parameter matrix of $v$-th view for the $i$-th instance, $s(\cdot)$ is a similarity function, and  $h_i^+$ is given by
\begin{equation}
    h_i^+ = \frac{1}{V}\sum_{v=1}^V h_i^{(v)}
\end{equation}
Then, common representation $\tilde{h}_i$ is obtained by the weighted average of features with attention distribution $\alpha_i^{(v)}$:
\begin{equation}
    \tilde{h}_i = \sum_{v=1}^V \alpha_i^{(v)} h_i^{(v)}
\end{equation}

After obtaining the common representation $\{\tilde{h}_i\}_{i=1}^N$, view-specific decoders $d^{(v)}$ is employed to reconstruct the original multi-view data. And the traditional element-wise mean squared error shown in Eq.(7) is utilized to evaluate reconstruction in the pre-training.
\begin{equation}
    L_{mr} = \sum_{v=1}^V \sum_{i=1}^N ||x_i^{(v)}- d^{(v)}(\tilde{h}_i)||^2
\end{equation}

\begin{algorithm}[!t]\small
\caption{The CRTC algorithm}
\begin{algorithmic}
\STATE \textbf{Input:} A multi-view incomplete dataset $\mathcal{X} = \{X^{(1)},...,X^{(V)}\}$, number of clusters: $C$, maximum iterations: $MaxIter$, stopping threshold: $\delta$, target distribution update interval: $T$.
\STATE \textbf{Output:} Clustering assignments $R$.
\STATE \textbf{Initialization:} Initialize parameters of the whole model, construct KNN graphs $K_a$ for missing data.
\STATE Pre-train cross-view relation transfer completion module using Eq.(2).
\STATE Pre-train multi-view fusion clustering network using Eq.(7).
\STATE Obtain cluster centers $\{\mu_j\}_{j=1}^C$ using K-means.
\STATE Obtain target distribution $P$ according to Eq.(10).
\FOR{$Iter=0$ \TO $MaxIter$}
\IF{$Iter\%T=0$}
\STATE Recover missing data according to Eq.(1).
\STATE Compute the common representation $\tilde{H}$.
\STATE Compute $P$ and $Q$ using Eq.(9) and Eq.(10).
\STATE Save last clustering assignments: $r_{old} = r$.
\STATE Compute clustering assignments $r$ via $r_i = \mathop{\arg\min}\limits_{j}q_{ij}$.
\IF{$sum(r_{old} \neq r)/N \leq \delta$}
\STATE Stop training.
\ENDIF
\ENDIF
\STATE Update cross-view relation transfer completion module by Eq.(11).
\STATE Update multi-view fusion clustering network by Eq.(8).
\ENDFOR
\RETURN Clustering assignments $R$.
\end{algorithmic}
\end{algorithm}

\begin{table*}[h!]
\small
\centering
\caption{Clustering results on five datasets with different missing-view rates or paired-view rates (Average$\pm$Standard deviations,$\%$). Note: the results of PVC-GAN and GP-MVC shown in the table are from \cite{journals/tip/WangDTGF21}.}
\resizebox{15cm}{!}{
\begin{tabular}{c|l|cccc|cccc}
\toprule
\multicolumn{2}{c|}{}                           & \multicolumn{4}{c|}{ACC (\%)} & \multicolumn{4}{c}{NMI (\%)} \\
\midrule
Datasets                    & Method\textbackslash{}$p$ & 0.1  & 0.3  & 0.5 & 0.7 & 0.1  & 0.3  & 0.5 & 0.7  \\
\midrule
\multirow{11}{*}{BDGP}      & BSV                       & 36.73$\pm$1.96 & 42.21$\pm$2.04 &	44.72$\pm$4.33 & 50.15$\pm$2.47 & 25.27$\pm$0.72 & 30.67$\pm$1.11 & 35.00$\pm$2.52 & 38.74$\pm$1.05 \\
                            & Concat                    & 40.66$\pm$1.60 & 45.74$\pm$1.53 &	48.46$\pm$2.23 & 55.41$\pm$2.61 & 28.57$\pm$1.18 & 33.58$\pm$1.58 &	36.50$\pm$1.97 & 43.28$\pm$2.48 \\
                            & PVC                       & 42.54$\pm$0.00 & 57.84$\pm$0.00 &	57.10$\pm$0.00 & 60.47$\pm$0.00 & 23.59$\pm$0.00 & 34.79$\pm$0.00 & 33.86$\pm$0.00 & 38.33$\pm$0.00 \\
                            & MIC                       & 34.22$\pm$1.66 & 41.23$\pm$2.62 &	51.55$\pm$1.90 & 58.78$\pm$4.57 & 13.33$\pm$2.00 & 20.31$\pm$2.56 & 30.76$\pm$0.50 & 35.82$\pm$3.48 \\
                            & IMG                       & 48.33$\pm$0.89 & 50.78$\pm$0.17 &	53.61$\pm$0.16 & 55.75$\pm$0.84 & 31.81$\pm$1.20 & 34.62$\pm$0.12 & 37.95$\pm$0.75 & 38.73$\pm$1.15 \\
                            & DAIMC                     & 50.34$\pm$6.38 & 57.22$\pm$5.22 &	62.92$\pm$5.20 & 67.49$\pm$1.14 & 25.05$\pm$4.86 & 34.30$\pm$3.67 &	44.43$\pm$2.16 & 50.21$\pm$0.89 \\
                            & PVC-GAN                   & 52.10$\pm$0.90 & 67.11$\pm$1.07 & 86.31$\pm$0.43 & 91.54$\pm$1.07 & -              & -              & -              & -              \\
                            & GP-MVC                    & 58.74$\pm$2.49 & 78.68$\pm$2.34 & 88.79$\pm$1.28 & 93.19$\pm$0.82 & -              & -              & -              & -              \\
                            & iCmSC                     & 53.04$\pm$2.18 & 67.90$\pm$2.09 &	80.27$\pm$0.74 & 89.63$\pm$0.98 & 39.27$\pm$1.38 & 44.63$\pm$2.64 & 61.28$\pm$1.32 & 64.92$\pm$0.95 \\
                            & CDIMC-net                 & 57.43$\pm$3.05 & 74.72$\pm$2.05 &	77.65$\pm$1.50 & 83.69$\pm$2.98 & 36.25$\pm$4.19 & 53.26$\pm$3.10 &	62.02$\pm$2.33 & 67.63$\pm$3.59 \\
                            & CRTC                      & \textbf{70.28$\pm$5.89} & \textbf{85.62$\pm$3.88} & \textbf{90.34$\pm$2.09} & \textbf{93.88$\pm$1.29} & \textbf{45.45$\pm$5.78} & \textbf{67.55$\pm$5.33} & \textbf{75.77$\pm$3.33} & \textbf{81.75$\pm$1.91} \\
\midrule
\multirow{11}{*}{MNIST}     & BSV                       & 34.03$\pm$0.97 & 39.79$\pm$1.96 &	43.24$\pm$1.37 & 49.15$\pm$1.65 & 30.00$\pm$0.86 & 33.76$\pm$1.25 &	37.00$\pm$0.95 & 41.32$\pm$1.00 \\
                            & Concat                    & 40.00$\pm$0.78 & 41.07$\pm$1.06 & 43.99$\pm$2.27 & 48.17$\pm$1.89 & 34.40$\pm$0.53 & 35.63$\pm$0.81 &	37.03$\pm$0.91 & 41.49$\pm$1.08 \\
                            & PVC                       & 39.88$\pm$0.00 & 44.20$\pm$0.00 &	46.97$\pm$0.00 & 44.43$\pm$0.00 & 35.03$\pm$0.00 & 37.34$\pm$0.00 & 39.59$\pm$0.00 & 39.75$\pm$0.00 \\
                            & MIC                       & 36.31$\pm$4.39 & 30.83$\pm$2.84 &	36.78$\pm$1.57 & 38.24$\pm$2.50 & 26.08$\pm$3.91 & 25.75$\pm$2.43 &	31.00$\pm$1.50 & 33.40$\pm$1.90 \\
                            & IMG                       & 43.15$\pm$1.03 & 43.38$\pm$0.24 &	46.20$\pm$0.98 & 48.08$\pm$1.46 & 33.83$\pm$0.38 & 36.03$\pm$0.11 & 36.58$\pm$0.43 & 38.68$\pm$0.19 \\
                            & DAIMC                     & 41.39$\pm$3.87 & 48.75$\pm$2.69 &	50.75$\pm$1.32 & 51.14$\pm$1.50 & 33.53$\pm$2.92 & 40.99$\pm$2.20 &	43.42$\pm$0.95 & 44.14$\pm$1.21 \\
                            & PVC-GAN                   & 45.17$\pm$0.86 & 48.36$\pm$0.71 & 52.80$\pm$0.78 & 52.02$\pm$0.70 & -              & -              & -              & -              \\
                            & GP-MVC                    & 56.46$\pm$2.47 & 55.42$\pm$3.30 & 57.76$\pm$1.69 & 59.63$\pm$0.88 & -              & -              & -              & -              \\
                            & iCmSC                     & 46.97$\pm$0.26 & 51.11$\pm$0.58 & 56.05$\pm$1.12 & 58.65$\pm$0.80 & 48.66$\pm$1.35 & 53.27$\pm$0.40 & 54.21$\pm$1.20 & 55.48$\pm$1.79 \\
                            & CDIMC-net                 & 51.65$\pm$0.14 & 57.64$\pm$1.44 & 58.28$\pm$0.68 & 59.15$\pm$0.21 & 48.25$\pm$0.47 & 50.54$\pm$1.26 & 51.70$\pm$0.67 & 52.87$\pm$0.48 \\
                            & CRTC                      & \textbf{58.27$\pm$0.36} & \textbf{58.64$\pm$0.91} & \textbf{59.88$\pm$1.48} & \textbf{61.02$\pm$0.59} & \textbf{52.89$\pm$1.10} & \textbf{55.75$\pm$0.82} & \textbf{56.04$\pm$1.94} & \textbf{56.80$\pm$0.99} \\
\midrule
\multirow{9}{*}{Caltech20}  & BSV                       & 32.80$\pm$0.65 & 33.62$\pm$0.94 &	33.03$\pm$1.95 & 36.59$\pm$1.55 & 36.27$\pm$0.46 & 42.00$\pm$0.60 &	45.05$\pm$1.11 & 49.67$\pm$0.78 \\
                            & Concat                    & 33.65$\pm$1.97 & 31.74$\pm$1.50 &	31.96$\pm$2.27 & 36.33$\pm$2.12 & 38.34$\pm$0.61 & 42.13$\pm$1.11 &	46.60$\pm$1.09 & 51.23$\pm$1.48 \\
                            & PVC                       & 36.43$\pm$0.00 & 36.34$\pm$0.00 & 41.75$\pm$0.00 & 44.25$\pm$0.00 & 50.12$\pm$0.00 & 50.65$\pm$0.00 &	53.43$\pm$0.00 & 57.10$\pm$0.00 \\
                            & MIC                       & 30.91$\pm$2.32 & 31.86$\pm$2.15 &	32.14$\pm$2.01 & 37.66$\pm$2.31 & 40.70$\pm$1.35 & 43.93$\pm$1.64 &	46.27$\pm$1.15 & 51.90$\pm$0.60 \\
                            & IMG                       & 39.37$\pm$0.83 & 42.65$\pm$0.24 &	43.93$\pm$0.57 & 44.91$\pm$0.26 & 49.56$\pm$1.28 & 53.27$\pm$0.27 & 54.16$\pm$0.26 & 56.08$\pm$0.22 \\
                            & DAIMC                     & 31.97$\pm$3.18 & 41.14$\pm$2.91 &	45.08$\pm$2.18 & 46.53$\pm$2.34 & 46.04$\pm$1.59 & 50.00$\pm$2.23 &	52.68$\pm$1.26 & 54.68$\pm$1.31 \\
                            & iCmSC                     & 46.81$\pm$0.59 & 45.88$\pm$1.17 &	51.90$\pm$1.68 & 53.06$\pm$1.84 & 60.05$\pm$0.91 & 59.49$\pm$0.47 &	61.80$\pm$1.59 & 64.15$\pm$1.34 \\
                            & CDIMC-net                 & 46.38$\pm$5.29 & 49.63$\pm$3.77 &	50.46$\pm$1.73 & 51.23$\pm$2.56 & 59.65$\pm$0.40 & 62.21$\pm$0.28 &	64.73$\pm$0.82 & 65.15$\pm$0.86 \\
                            & CRTC                      & \textbf{61.96$\pm$2.75} & \textbf{63.83$\pm$2.04} & \textbf{65.33$\pm$3.15} & \textbf{66.71$\pm$4.09} & \textbf{62.11$\pm$3.65} & \textbf{63.90$\pm$2.26} & \textbf{64.87$\pm$3.01} & \textbf{66.10$\pm$3.26} \\
\midrule
\multirow{8}{*}{Animal}     & BSV                       & 35.64$\pm$0.90 & 43.39$\pm$1.28 &	49.11$\pm$1.04 & 55.62$\pm$1.13 & 49.03$\pm$0.20 & 55.69$\pm$0.51 &	61.10$\pm$0.47 & 67.16$\pm$0.45 \\
                            & Concat                    & 38.69$\pm$1.14 & 43.71$\pm$1.24 & 49.05$\pm$1.67 & 54.65$\pm$1.65 & 52.61$\pm$0.52 & 56.50$\pm$0.42 &	60.49$\pm$0.58 & 64.85$\pm$0.50 \\
                            & MIC                       & 34.26$\pm$0.93 & 39.02$\pm$1.79 & 41.29$\pm$0.89 & 48.45$\pm$1.87 & 48.36$\pm$1.14 & 51.22$\pm$1.37 &	55.11$\pm$0.89 & 59.88$\pm$0.59 \\
                            & IMG                       & \textbf{48.91$\pm$0.11} & 54.52$\pm$0.23 & 56.81$\pm$0.19 & 59.67$\pm$0.08  & 54.16$\pm$0.08 & 58.58$\pm$0.08 &	61.40$\pm$0.10 & 65.27$\pm$0.13 \\
                            & DAIMC                     & 41.45$\pm$2.22 & 51.51$\pm$1.00 &	55.65$\pm$1.13 & 57.36$\pm$0.84 & 49.45$\pm$1.18 & 56.34$\pm$0.76 &	60.72$\pm$0.42 & 64.34$\pm$0.35 \\
                            & iCmSC                     & 37.40$\pm$1.58 & 40.27$\pm$1.01 & 44.31$\pm$4.00 & 47.71$\pm$1.20 & 52.82$\pm$1.01 & 54.74$\pm$1.51 & 55.73$\pm$0.70 & 59.45$\pm$0.93 \\
                            & CDIMC-net                 & 36.29$\pm$0.67 & 45.43$\pm$1.37 &	49.61$\pm$1.44 & 54.80$\pm$1.56 & 51.55$\pm$0.25 & 56.52$\pm$0.27 &	60.03$\pm$0.66 & 63.46$\pm$0.68 \\
                            & CRTC                      & 44.49$\pm$1.13 & \textbf{55.02$\pm$1.69} & \textbf{62.41$\pm$1.03} & \textbf{66.63$\pm$1.76} & \textbf{54.56$\pm$0.15} & \textbf{60.01$\pm$0.36} & \textbf{68.25$\pm$0.52} & \textbf{72.51$\pm$0.69} \\
\midrule
\multirow{8}{*}{HW}         & BSV                       & 69.05$\pm$4.10 & 53.00$\pm$2.36 &	37.67$\pm$1.04 & 28.04$\pm$0.57 & 63.69$\pm$1.55 & 50.12$\pm$0.41 &	38.17$\pm$1.15 & 28.00$\pm$0.78 \\
                            & Concat                    & 57.10$\pm$2.67 & 46.49$\pm$2.35 & 35.58$\pm$2.05 & 26.52$\pm$0.10 & 54.36$\pm$1.29 & 43.98$\pm$1.61 &	33.83$\pm$1.16 & 24.75$\pm$0.15 \\
                            & PVC                       & 62.15$\pm$0.00 & 59.50$\pm$0.00 & 42.93$\pm$0.00 & 33.86$\pm$0.00 & 55.73$\pm$0.00 & 57.28$\pm$0.00 &	36.79$\pm$0.00 & 31.06$\pm$0.00 \\
                            & MIC                       & 70.05$\pm$1.05 & 68.43$\pm$1.96 &	59.87$\pm$0.59 & 38.45$\pm$3.12 & 68.89$\pm$1.82 & 65.31$\pm$2.53 &	50.12$\pm$1.31 & 30.69$\pm$2.56 \\
                            & IMG                       & 73.07$\pm$6.45 & 66.44$\pm$5.02 &	57.91$\pm$2.72 & 42.63$\pm$0.09 & 67.04$\pm$3.95 & 58.58$\pm$2.96 &	50.30$\pm$0.74 & 39.78$\pm$0.29 \\
                            & DAIMC                     & 83.29$\pm$4.82 & 82.22$\pm$3.72 &	75.86$\pm$5.72 & 49.51$\pm$6.14 & 76.19$\pm$3.21 & 73.58$\pm$2.80 &	66.01$\pm$3.29 & 40.35$\pm$5.81 \\
                            & CDIMC-net                 & 95.13$\pm$0.39 & 94.67$\pm$0.77 &	91.68$\pm$1.24 & 87.98$\pm$0.39 & 91.33$\pm$0.75 & 89.85$\pm$1.09 &	84.84$\pm$1.76 & \textbf{82.13$\pm$0.42} \\
                            & CRTC                      & \textbf{97.31$\pm$0.71} & \textbf{96.80$\pm$0.69} & \textbf{95.18$\pm$1.15} & \textbf{90.24$\pm$1.87} & \textbf{93.20$\pm$0.99} & \textbf{92.63$\pm$0.71} & \textbf{89.47$\pm$0.93} & 81.89$\pm$2.03 \\
\bottomrule
\end{tabular}}
\end{table*}

\subsection{Joint Training and Clustering}

By optimizing the above Eq.(2) and Eq.(7), we can obtain the recovery data and common representation $\tilde{H}= \{\tilde{h}_1,...,\tilde{h}_N\}$. However, there may be several potential mistakes in the transferred relations as the distinctness between views. Meanwhile, whether the recovery data and the common representation are suitable for clustering is unidentified. 
For this reason, we develop a dual self-supervised training module as a solution to confront this challenge in both modules. It can optimize recovery and clustering simultaneously to obtain the clustering assignments.

Specifically, for the common representation $\tilde{H}$, if its cluster center is denoted by $\mu_j$, the clustering loss function for the multi-view fusion clustering network can be written as 
\begin{equation}
L_{mc} = KL(P||Q) = \sum_{i=1}^N\sum_{j=1}^C p_{ij}\rm{log}\frac{\it p_{ij}}{\it q_{ij}}
\end{equation}
In Eq.(8), $q_{ij}$ can be calculated using the Student’s t-distribution\cite{jmlr/laurens08} as 
\begin{equation}
q_{ij} = \frac{(1+||\tilde{h}_i - \mu_j||^2)^{-1}}{\sum_{k=1}^C(1+||\tilde{h}_i - \mu_k||^2)^{-1}}
\end{equation}
where $\mu_j$ is initialized by K-means \cite{kmeans} on the pre-trained common representation. $q_{ij}$ can be seen as the probability of assigning node $\tilde{h}_i$ to cluster $j$. $Q = [q_{ij}]$ is the distribution of the assignments of all nodes. $P$ is the target distribution of $Q$ and can be calculated as 
\begin{equation}
p_{ij} = \frac{q_{ij}^2/\sum_{i=1}^N q_{ij}}{\sum_{k=1}^C(q_{ik}^2/\sum_{i=1}^N q_{ik})}
\end{equation}
By minimizing the KL divergence loss between distributions $P$ and $Q$, a more cluster-friendly representation can be produced according to the high confidence predictions.

To obtain more cluster-friendly recovery data and reduce the impact of possible errors in the transferred relations, KL divergence between the recovery data and relational instances is used to train the cross-view relation transfer completion module. And the loss function can be written as
\begin{equation}
L_{cc} = KL(Q_a||Q_K) = \sum_{i=1}^{N_a}\sum_{j \in K_{a}}\sum_{k=1}^C q_{ik}\rm{log}\frac{\it q_{ik}}{\it q_{ijk}}
\end{equation}
where $N_a$ is the number of missing instances. By optimizing Eq.(11), the clustering information can be introduced into the missing data recovery process to obtain more clustering-friendly completion features. Our CRTC method is summarized in Algorithm 1.

\begin{figure*}[!t]
	\centering
	\subfigure[BSV]{	
		\label{fig2:a} 
		\includegraphics[width=4.1cm]{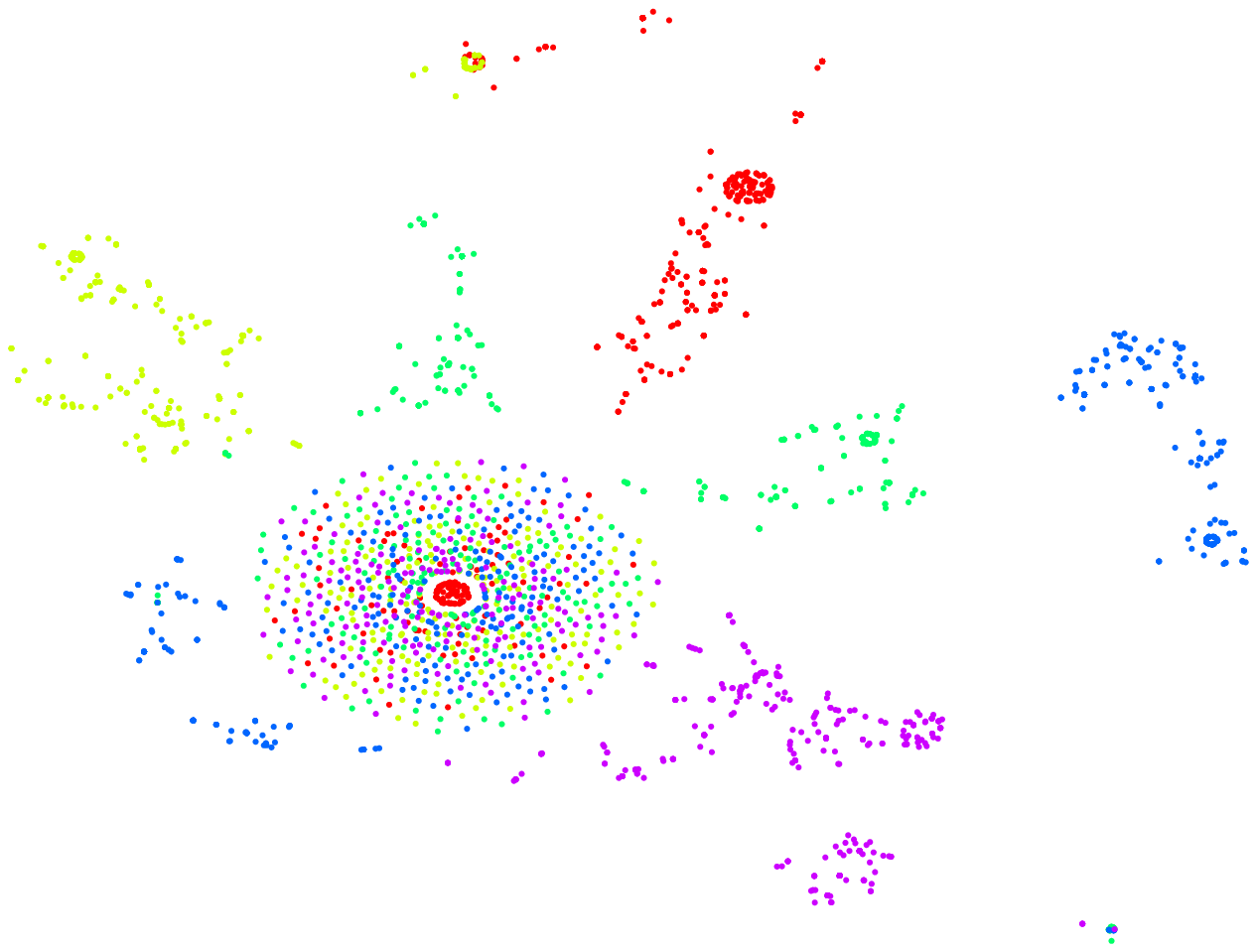}}
	\subfigure[Concat]{	
		\label{fig2:b}
		\includegraphics[width=4.1cm]{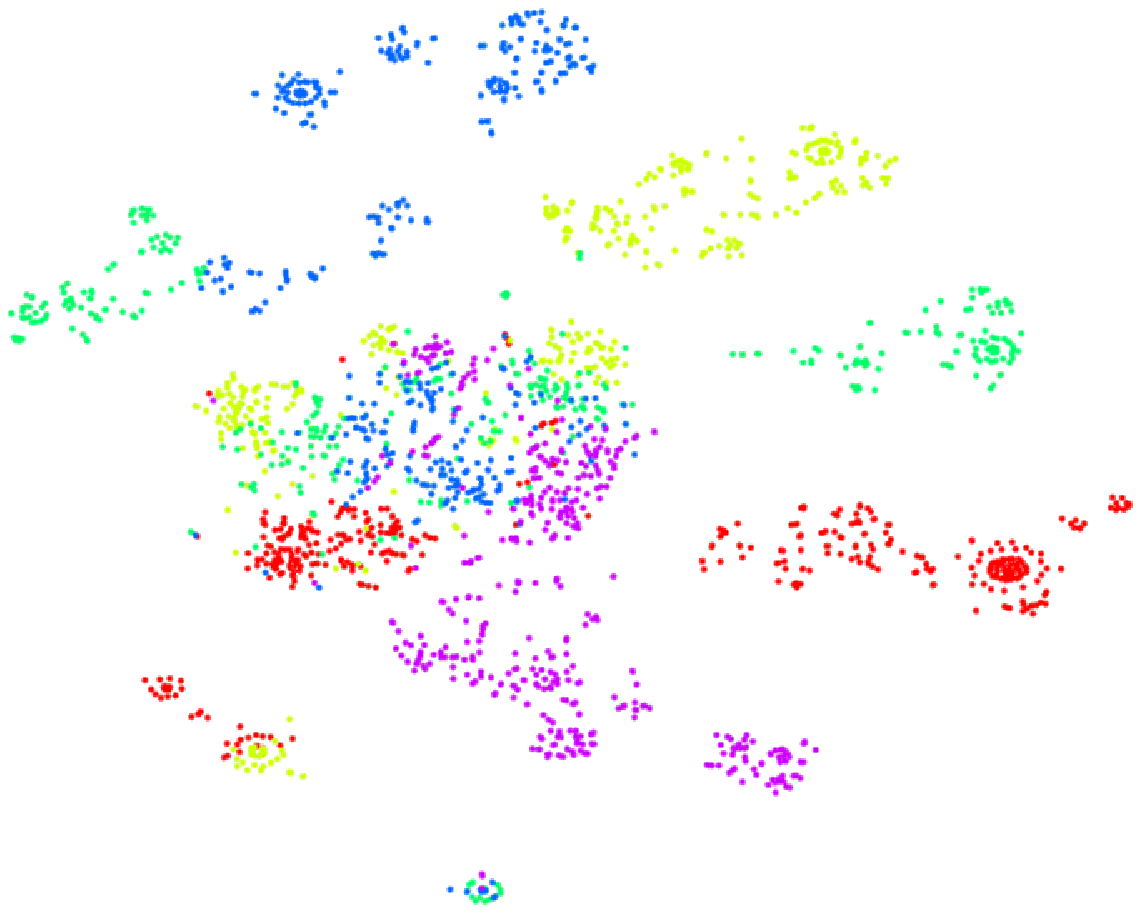}}
	\subfigure[IMG]{	
		\label{fig2:c} 
		\includegraphics[width=4.1cm]{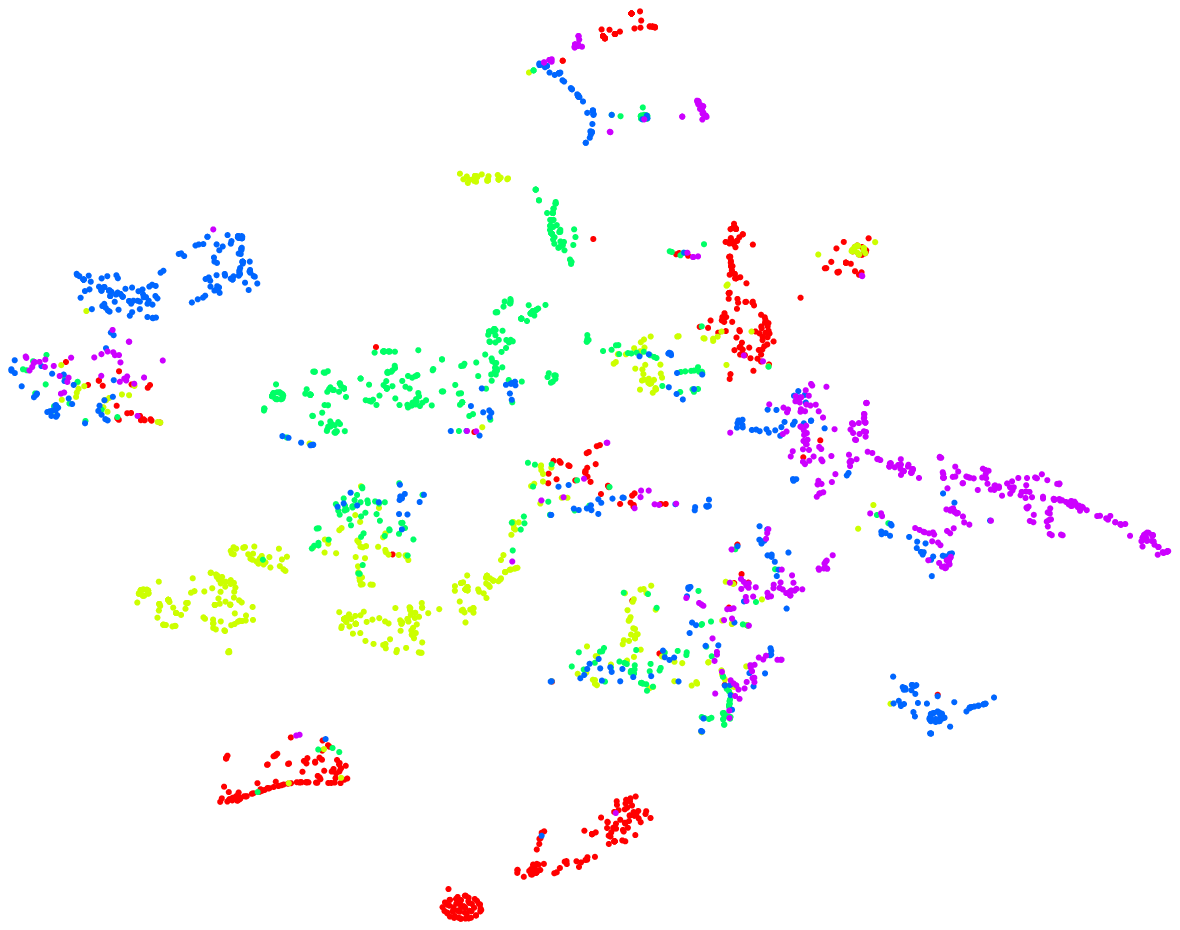}}
	\subfigure[MIC]{	
		\label{fig2:d}
		\includegraphics[width=4.1cm]{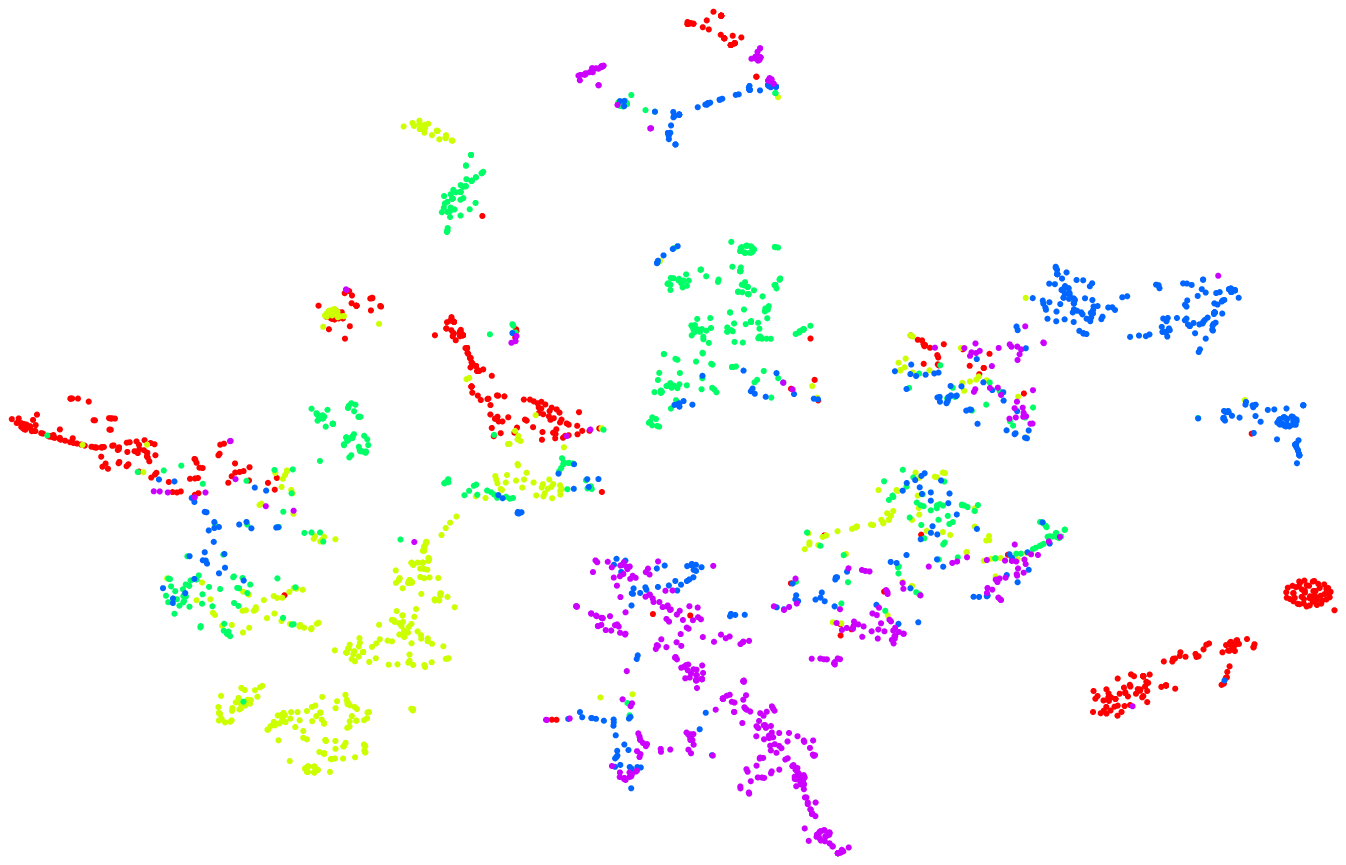}}
	\vfill
	\subfigure[DAIMC]{
		\label{fig2:e} 
		\includegraphics[width=4.1cm]{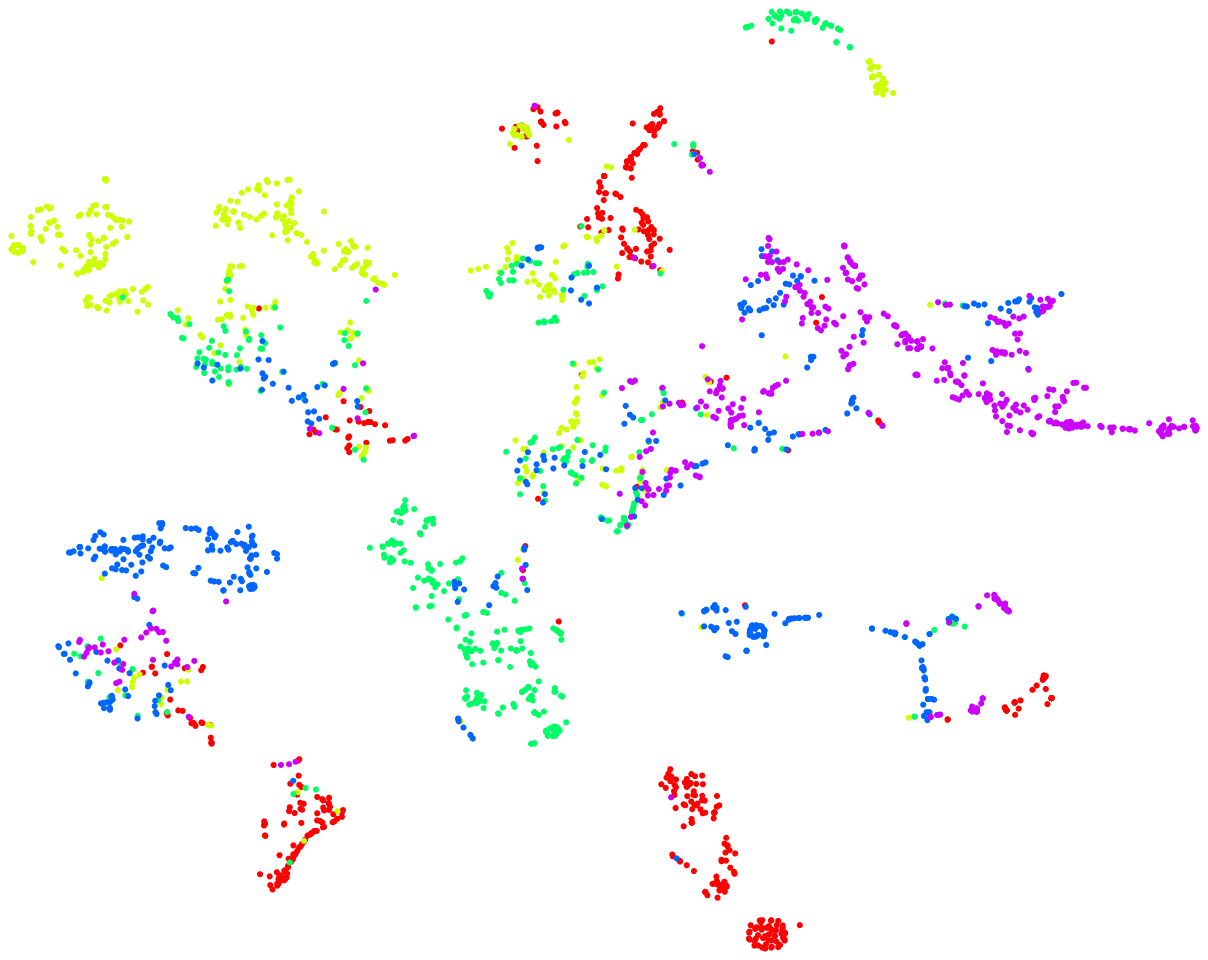}}
	\subfigure[CDIMC-net]{	
		\label{fig2:f} 
		\includegraphics[width=4.1cm]{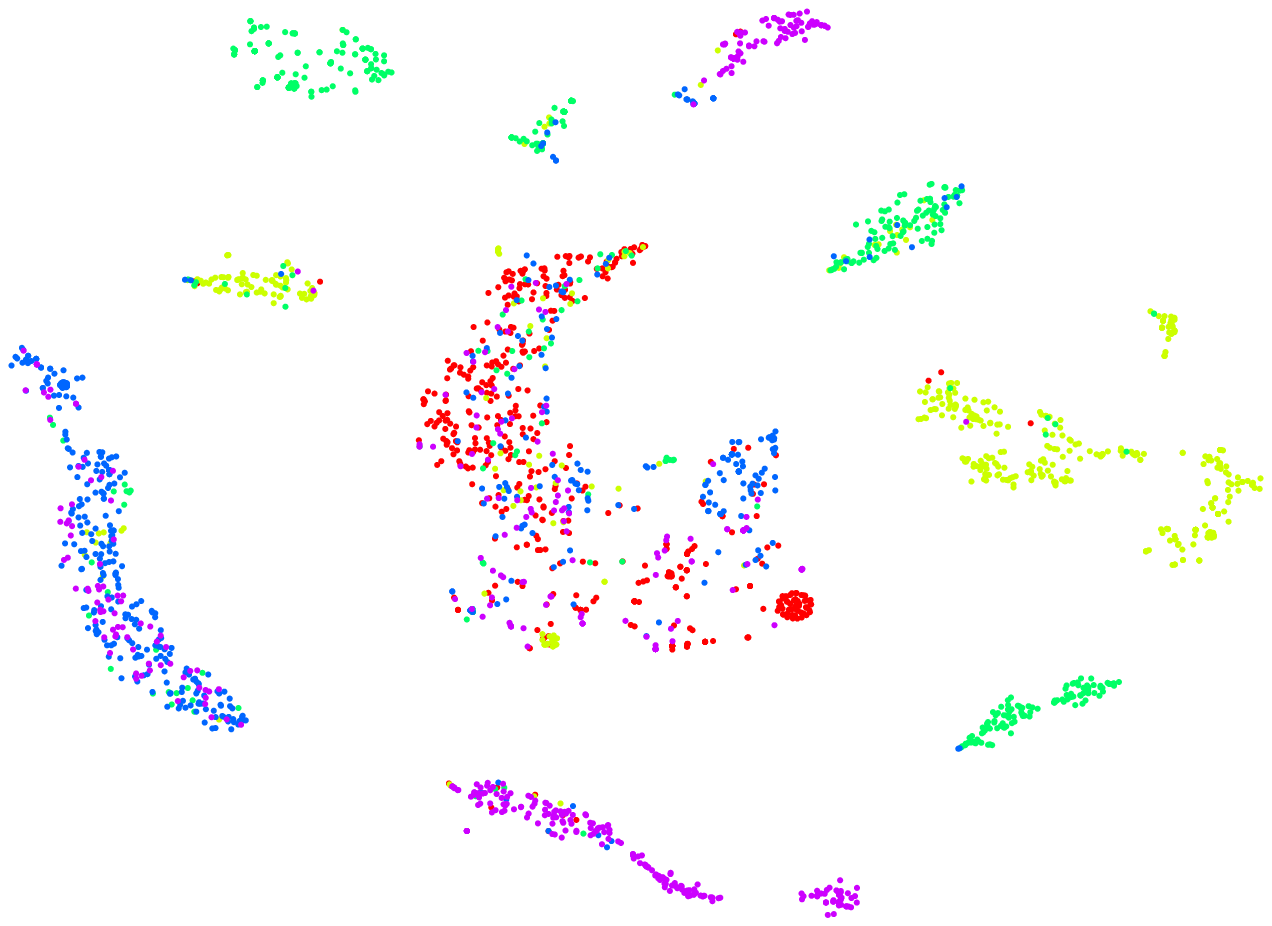}}
	\subfigure[CRTC-$h$]{	
		\label{fig2:g}
		\includegraphics[width=4.1cm]{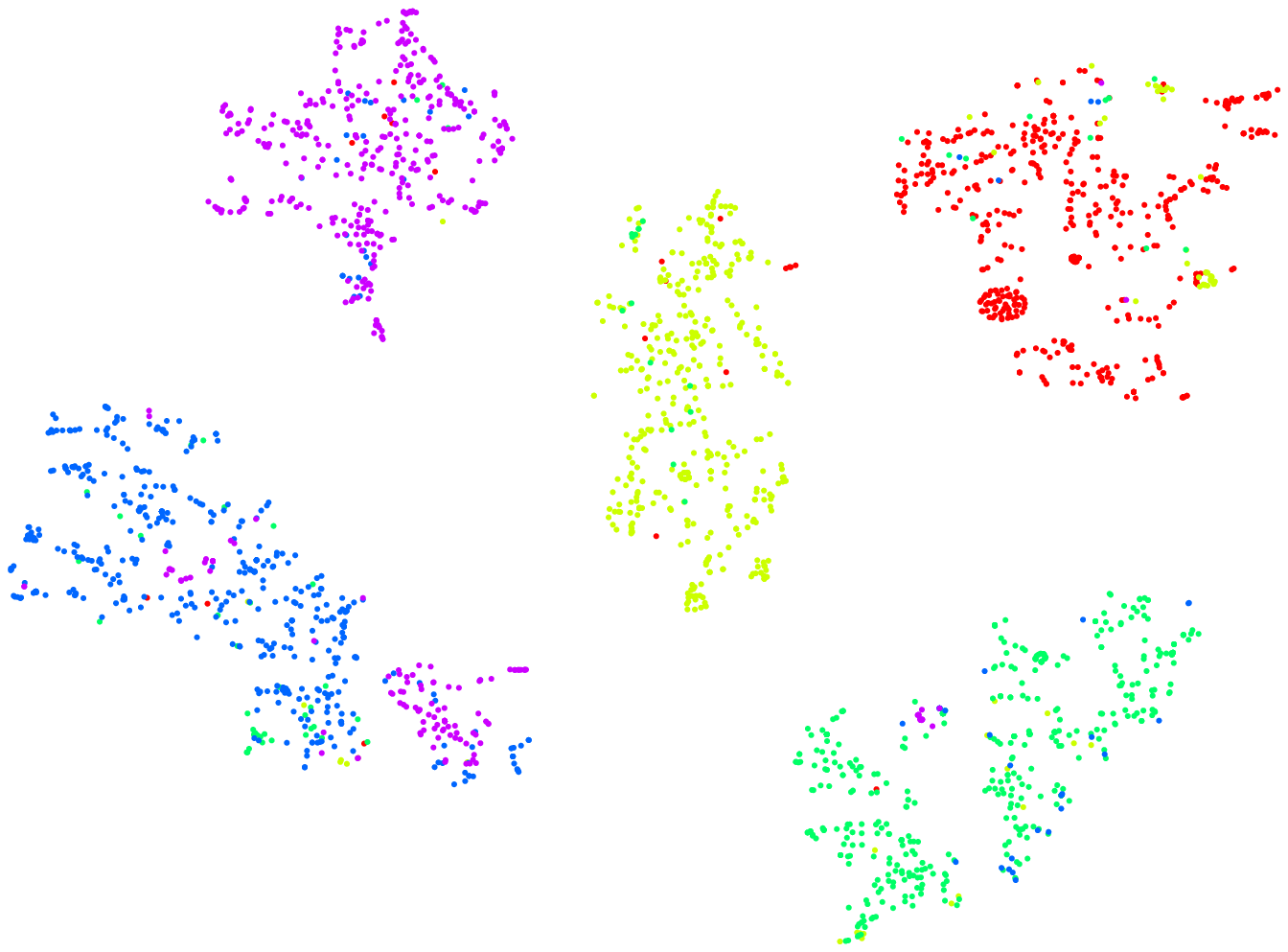}}
	\subfigure[CRTC-$q$]{	
		\label{fig2:h}
		\includegraphics[width=4.1cm]{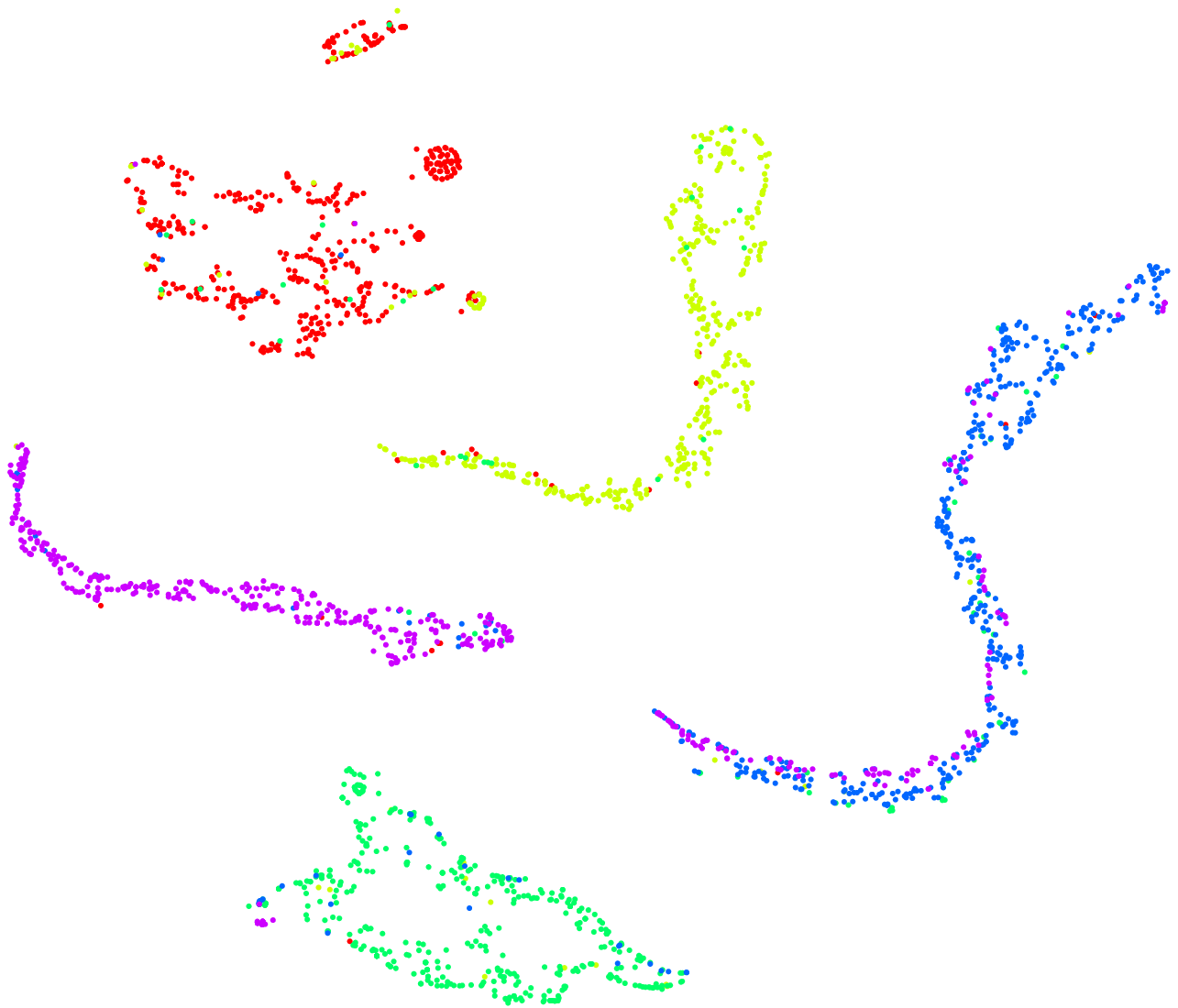}}
	\caption{$t$-SNE visualizations of representations learned by various methods on BDGP with a missing ratio of 30\%.}
	\label{fig2} 
\end{figure*}

\begin{figure}[t]
	\centering
    	\includegraphics[width=8.4cm]{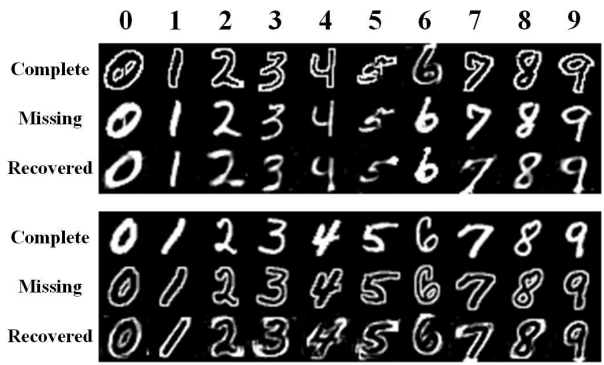}
	\caption{Data recovery on MNIST dataset.}
\end{figure}

\section{Experiments}
In this section, we evaluate the proposed CRTC method on five widely-used multi-view datasets with several state-of-the-art clustering methods.
\subsection{Experimental Settings}

\textbf{Datasets:}
Five widely-used datasets are used to assess the proposed CRTC method. 
1) \textbf{BDGP} \cite{journals/bioinformatics/CaiWHD12} contains 2500 images of five categories, and each image is described by a 1750-D visual vector and a 79-D textual feature vector.
2) \textbf{MNIST} is a widely-used dataset composed of 70000 digital images.
Since some baselines cannot handle such a large dataset, we follow \cite{conf/ijcai/WenZ0ZFX20, conf/icdm/WangDTG018} to use a subset of MNIST, which contains 4000 samples from 10 categories. Pixel feature and edge feature are extracted as two views. Please refer to the supplementary material for the results of the full MNIST dataset.
3) \textbf{Caltech20} \cite{conf/aaai/LiNHH15} consists of 2386 images of 20 subjects, and we follow \cite{lin2021completer} to use two views, \emph{i.e.} , HOG and GIST features. 
4) \textbf{Animal} consists of 10,158 images from 50 classes, and two types of deep features extracted with DECAF \cite{conf/nips/KrizhevskySH12} and VGG19 \cite{journals/corr/SimonyanZ14a} are used as two views. 
5) \textbf{HW} contains 2000 instances from ten numerals. We use five views from Fourier coefficients, profile correlations, KL coefficient, Zernike moments, and pixel average extractors. 

\textbf{Compared methods and evaluate metrics:} We compare CRTC with several baselines include: best single view K-means++ (BSV) \cite{conf/ijcai/ZhaoLF16}, concat feature K-means++ (Concat) \cite{conf/ijcai/ZhaoLF16}, PVC \cite{li2014partial}, MIC \cite{conf/pkdd/ShaoHY15}, IMG \cite{conf/dicta/QianSGTD16},  DAIMC \cite{conf/ijcai/HuC18}, PVC-GAN \cite{conf/icdm/WangDTG018}, CDIMC-net \cite{conf/ijcai/WenZ0ZFX20}, GP-MVC \cite{journals/tip/WangDTGF21}, and iCmSC \cite{journals/tip/WangLSGJ21}. And we evaluate the clustering performance with three statistical metrics:  Accuracy (ACC), Normalized Mutual Information (NMI), and Adjusted Rand Index (ARI). For all the metrics, higher value indicates a better performance. Please refer to the supplementary material for the results of ARI. 

\textbf{Implementation details:} For datasets with more than two views, we randomly remove $p(p\in\{0.1, 0.3, 0.5, 0.7\})$ instances under the condition that all samples at least have one view. For datasets with two views, we randomly select $p(p\in\{0.1, 0.3, 0.5, 0.7\})$ instances as paired samples whose views are complete, and the remaining samples are treated as single view samples. To better illustrate the effectiveness of the algorithm, we perform ten times and report the average clustering results and the corresponding standard deviation of all the methods. Note that our model is not sensitive to the $k$ of initial KNN graphs in a large range. We set $k=10$ for datasets with two views and set $k=5$ for the HW dataset. Please refer to the supplementary material for the details of network architectures and parameter settings.

\subsection{Performance with Different Missing Rates}

Table 1 shows the clustering performance of CRTC and baseline methods. The best results are highlighted in bold. Here, we summarize some observations as follows: (1) In terms of both ACC and NMI, CRTC achieves relatively promising performance compared with all baselines, which validates the effectiveness of the proposed method. (2) We can observe that incomplete multi-view methods achieve better results than BSV and Concat in most cases, which illustrates that exploring features from multiple sources can improve clustering performance. (3) CRTC achieves much better performance than GAN-based methods GP-MVC and PVC-GAN, validating that recovering missing data based on relation transfer can improve the performance of incomplete multi-view clustering.

We visualize the results of different methods on the BDGP dataset with a missing ratio of 30\% with $t$-SNE\cite{jmlr/laurens08}. The results are shown in Fig.~3, where CRTC-$h$ and CRTC-$q$ denote the common representation $\tilde{H}$ and the probability $Q$, respectively. As can be seen from Fig.~3, our method can recover a better cluster structure of data since it has smaller intra-cluster scatter and larger inter-cluster scatter, which demonstrates the effectiveness of the proposed CRTC method. Meanwhile, we can find that CRTC-$q$ has fewer intra-cluster errors and smaller intra-cluster scatter than CRTC-$h$, which illustrates that the clustering layer is valid for learning a better clustering structure.

\subsection{Model Analysis}

To visually present the recovered results of our method, we show the recovered images by our method on the MNIST dataset with a missing ratio of 30$\%$ on Fig.~4, in which lines 1 and 4 are complete views, lines 2 and 5 are missing views, and lines 3 and 6 are the recovered results from the complete view. We can see our method can recover the missing data for both the original and edge images, which illustrates that recovering missing data based on the similar relationships of existing instances is effective.

\begin{figure}[t]
	\centering
	\includegraphics[width=8.4cm]{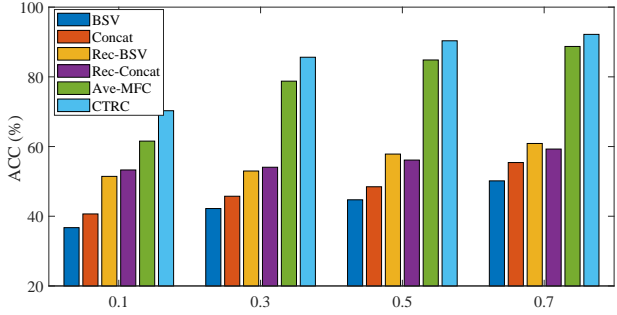}
	\caption{Performance comparisons of different approaches on BDGP dataset.}
\end{figure}

We further conduct a series of experiments with different approaches to show the effectiveness of the cross-view relation transfer completion. Fig.~5 shows the clustering accuracy on the BDGP dataset for six approaches: (a) BSV, (b) Concat, (c) BSV based on recovered data using the cross-view relation transfer (Rec-BSV), (d) Concat based on recovered data using the cross-view relation transfer (Rec-Concat), (e) Filling the missing data using the average value of the existing data and clustering by multi-view fusion clustering (Ave-MFC), and (f) the complete CRTC. According to Fig.~5, we have the following observations: (1) Compared with BSV and Concat, recovering data using the cross-view relation transfer can substantially improve the performance of k-means++ \cite{kmeans} for about $14\%$. The improvement is more significant with the decrease of paired-view rates. (2) The complete CRTC outperforms Ave-MFC in all paired-view rates, especially with less complete data, which demonstrates the validity of recovering data using the cross-view relation transfer.

\begin{figure}[!t]
	\centering
		\subfigure[ACC]{	
		\label{fige:a}
		\includegraphics[width=4.2cm]{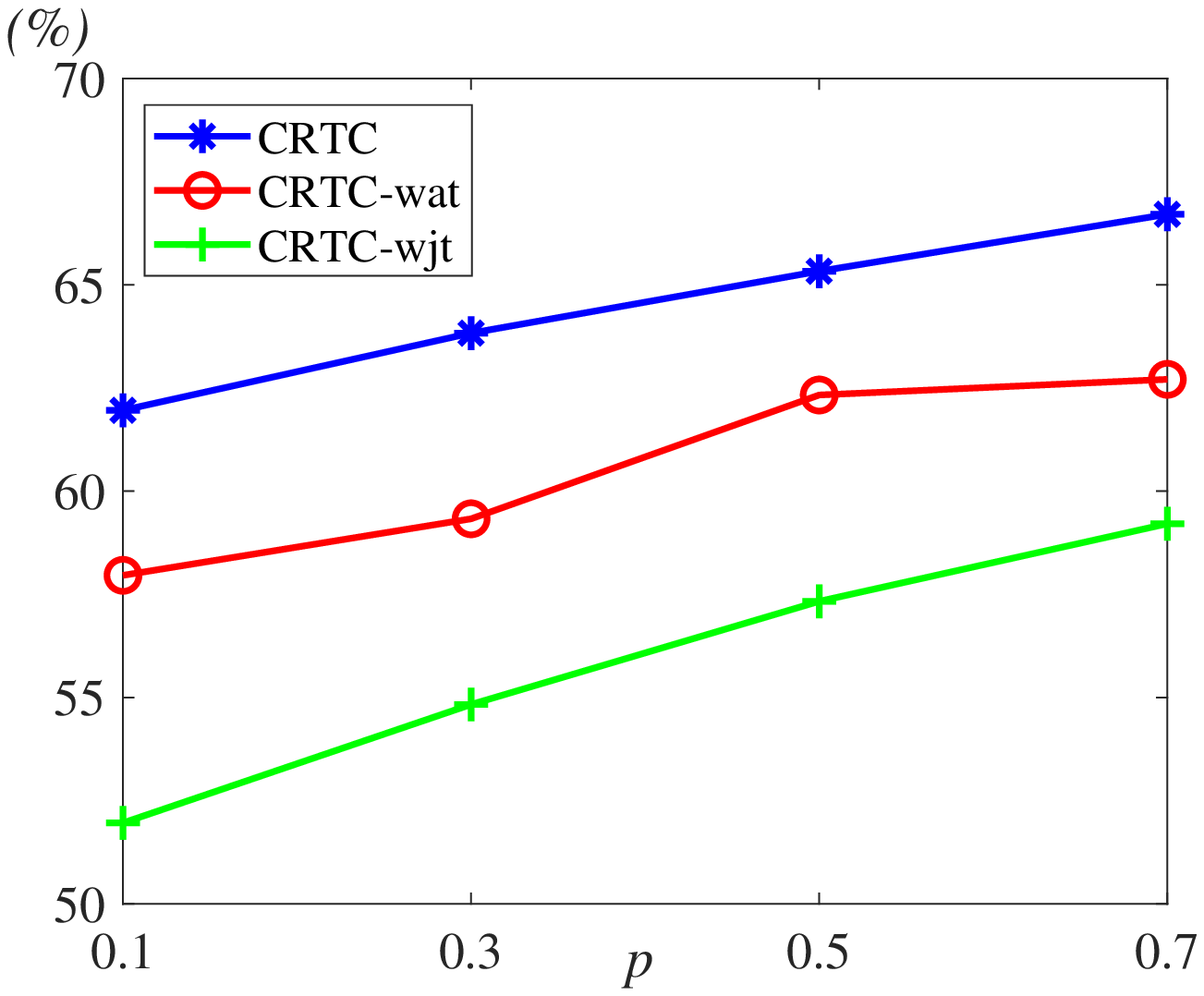}}\hspace{-3mm}
	\subfigure[NMI]{
		\label{fige:b} 
		\includegraphics[width=4.2cm]{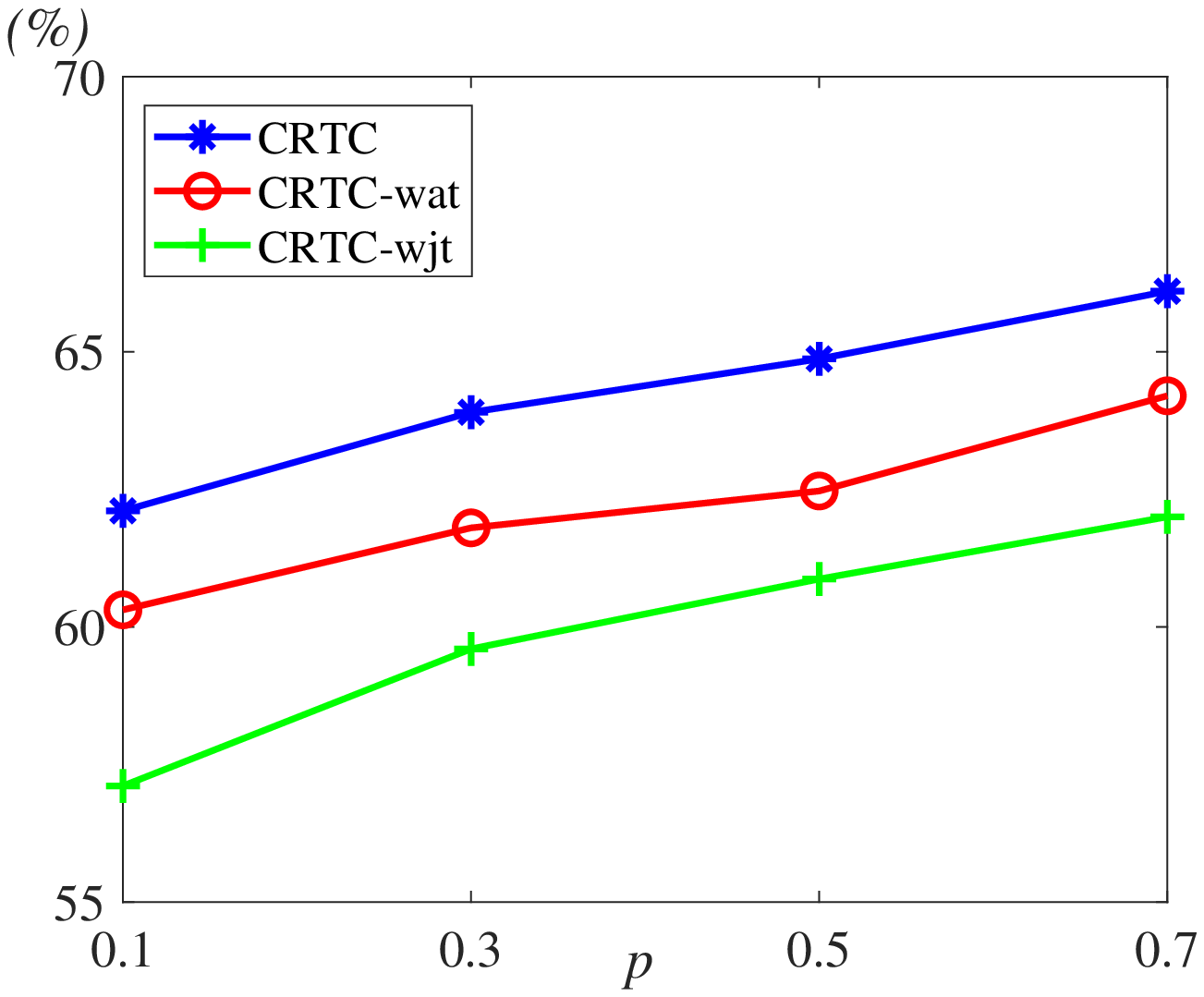}}
	\caption{ACC and NMI of the proposed CRTC and two degraded approaches on Caltech20.}
	\label{fig3} 
\end{figure}

In order to show the effect of the attention fusion layer and joint training, two degraded approaches of CRTC, \emph{i.e.}, CRTC without joint training (CRTC-wjd) and CRTC without attention fusion layer (CRTC-waf), are compared with the complete CRTC. The ACC and NMI results on the Caltech20 dataset are shown in Fig.~6. It can be seen that CRTC vastly outperforms these two degraded approaches in all paired-view rates, which illustrates that the joint training and the attention fusion layer can effectively improve the clustering performance.

\begin{figure}[!t]
	\centering
		\subfigure[$p = 0.3$]{	
		\label{fig4:a}
		\includegraphics[width=4.2cm]{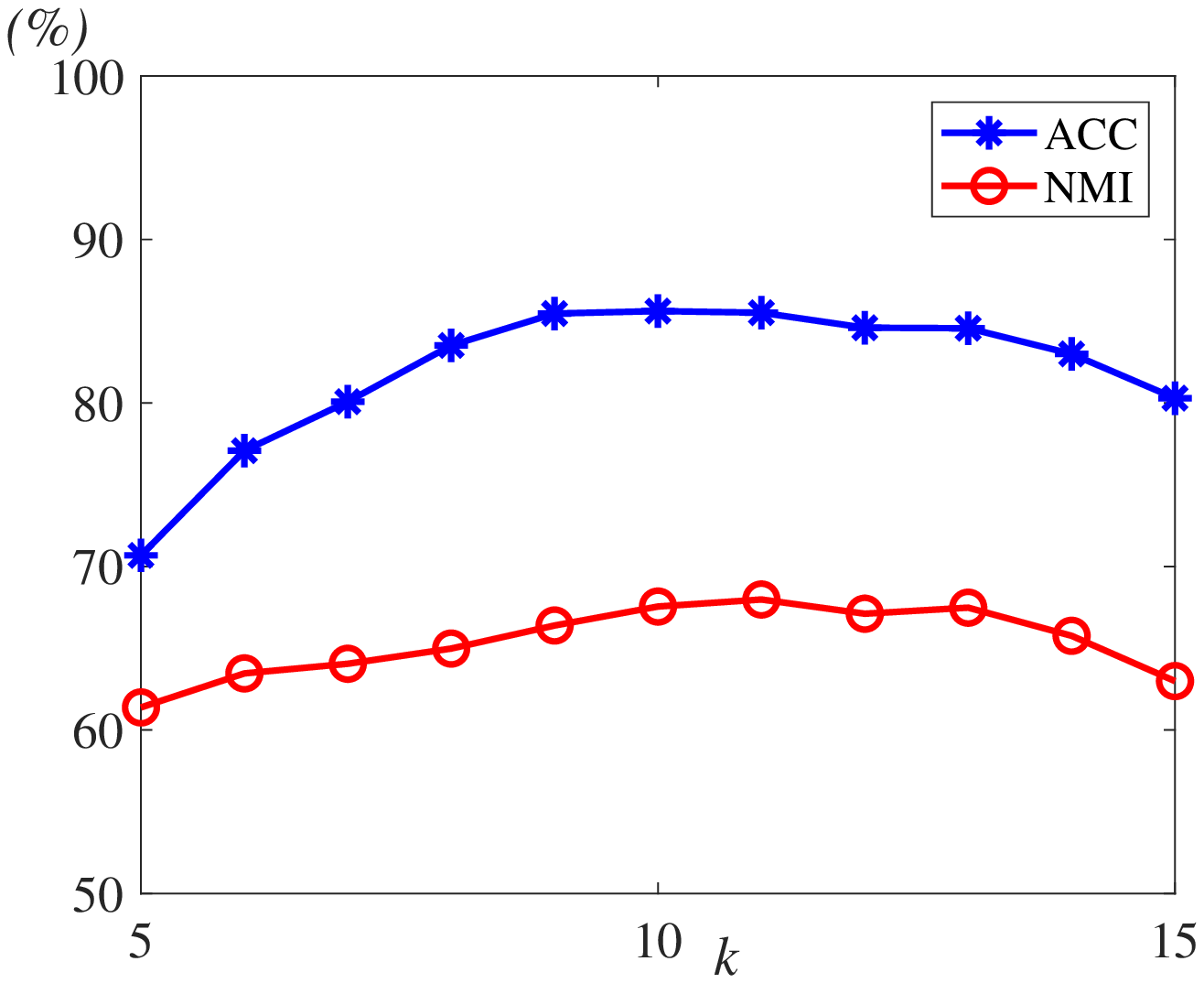}}\hspace{-3mm}
	    \subfigure[$p = 0.5$]{
		\label{fig4:b} 
		\includegraphics[width=4.2cm]{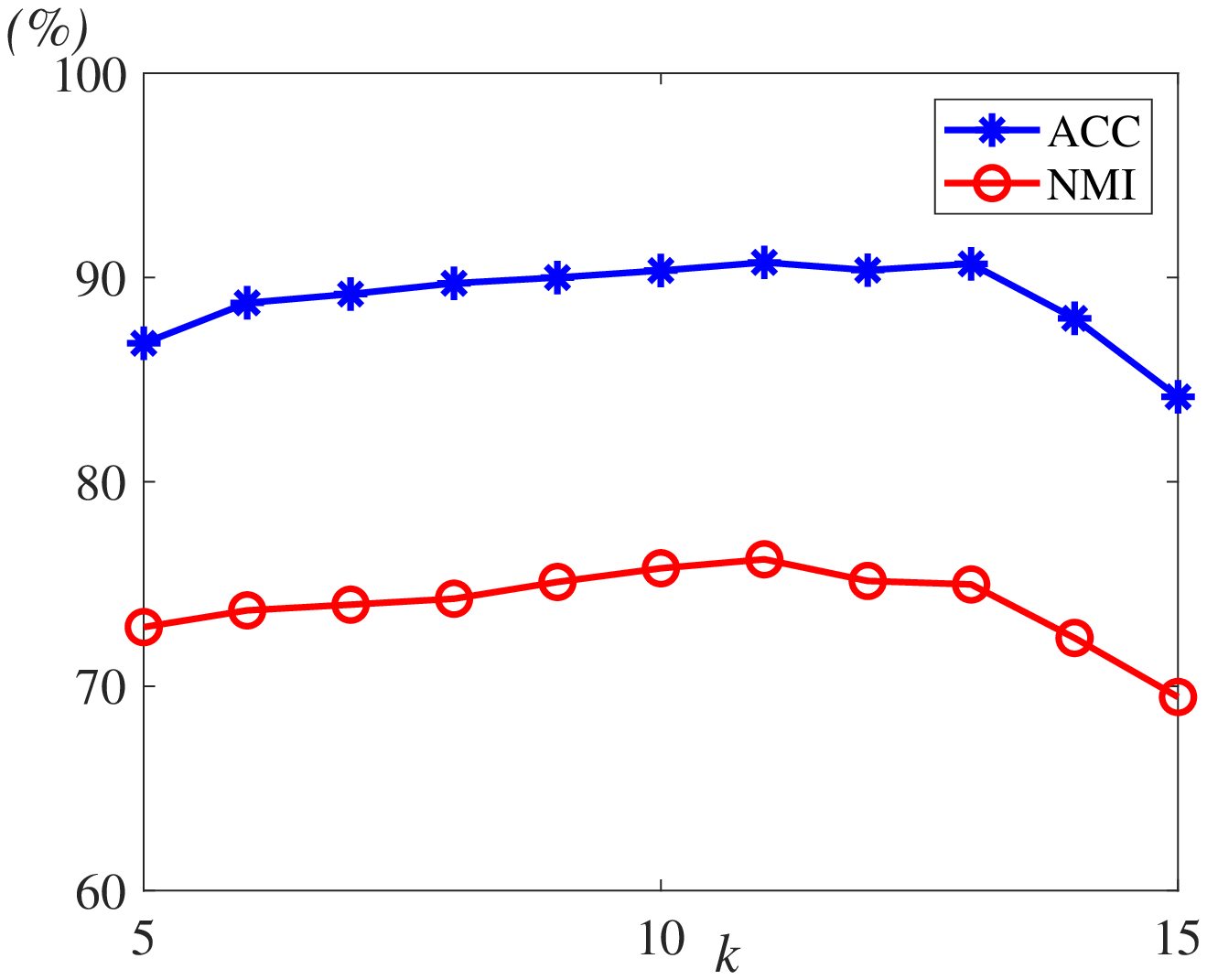}}
	\caption{The clustering performance influence of parameter $k$ on BDGP dataset with a missing ratio of 30$\%$.}
	\label{fig4} 
\end{figure}

The number of the nearest neighbors $k$ is an important parameter in constructing the $k$-NN graph and has a great impact on the performance of most graph-based algorithms. To examine the effect of $k$, we design a $k$-sensitivity experiment on the BDGP datasets with $k \in \{5,15\}$. It can be seen from Fig.~7 that our model is insensitive with $k$ in a larger range. It proves that our method is not sensitive to $k$ and can obtain a stable clustering even when there are fewer relational instances or some spurious relations in the transferred relation graph.

\section{Conclusion}
In this paper, we propose a novel algorithm for incomplete multi-view clustering named CRTC, which can simultaneously learn a clustering structure and recover the incomplete views. Based on the consensus that consistency exists between views, we construct similarity relations in the known views and transfer them to the missing views to recover incomplete data. Then, to fully explore the information in the transferred relations, GCN is employed to recover missing data based on the transferred relation graph and the existing instances. Compared with existing GAN-based approaches, the recovered views can fully exploit the relationships in existing views and are consistent with the clustering structure. Then, a multi-view fusion clustering network incorporating an attention-based fusion layer is introduced to obtain the clustering assignments. Moreover, a clustering layer is designed to optimize recovery and clustering simultaneously using the cluster-based KL-divergence loss. Comprehensive experiments validate the performance improvement of the proposed CRTC compared with state-of-the-art methods.

\bibliography{mybib}

\end{document}